\pdfoutput = 1

\documentclass[epj]{svjour}
\usepackage{graphics}
\usepackage{hyperref}
\usepackage{ulem}
\usepackage{multirow}

\usepackage{color}
\definecolor{greencolor}{rgb}{0,0.5,0.2}
\definecolor{redcolor}{rgb}{0.,0.,0.}
\definecolor{bluecolor}{rgb}{0,0.,1.}
\definecolor{greycolor}{rgb}{.5,.5,.5}
\def\Red#1{{\color{redcolor} #1}}

\begin{document}
\title{Discriminating word senses with tourist walks in complex networks}
\author{
Thiago C. Silva\inst{1} \and Diego R. Amancio\inst{2}% etc
}                     % Do not remove
\institute{
Institute of Mathematics and Computer Science \\
               University of S\~ao Paulo, P. O. Box 369, Postal Code 13560-970 \\
	           S\~ao Carlos, S\~ao Paulo, Brazil\\
\and
Institute of Physics of S\~ao Carlos \\
	University of S\~ao Paulo, P. O. Box 369, Postal Code 13560-970 \\
	S\~ao Carlos, S\~ao Paulo, Brazil \\
}
\date{Received: date / Revised version: date}
\abstract{
Patterns of topological arrangement are widely used for both animal and human brains in the learning process. Nevertheless, automatic learning techniques frequently overlook these patterns. In this paper, we apply a learning technique based on the structural organization of the data in the attribute space to the problem of discriminating the senses of $10$ polysemous words. Using two types of characterization of meanings, namely semantical and topological approaches, we have observed significative accuracy rates in identifying the suitable meanings in both techniques. Most importantly, we have found that the characterization based on the deterministic tourist walk improves the disambiguation process when one compares with the discrimination achieved with traditional complex networks measurements such as assortativity and clustering coefficient. To our knowledge, this is the first time that such deterministic walk has been applied to such a kind of problem. Therefore, our finding suggests that the tourist walk characterization may be useful in other related applications.
\PACS{
      {89.75.Hc}{Networks and genealogical trees}   \and
      {02.40.Pc}{General topology} \and
      {02.50.-r}{Probability theory, stochastic processes, and statistics}
     } % end of PACS codes
} %end of abstract
\maketitle
\section{Introduction}
\label{intro}

The last few years have seen a wave of automatic approaches devoted to processing natural languages~\cite{manning}. Phys\-i\-cists usually study language employing concepts from information theory~\cite{shanon} and other physical concepts~\cite{carpenas,carpenas2,ep1,ep2,ep3}.
More recently, language has been studied from a different standpoint as it can be schemed as a complex system. %~\cite{cs}.
Since traditional computational linguistic tools can hardly capture long-range information concerning the relationships between concepts, researchers have used the methods of complex networks~\cite{cn1} and non-linear systems to study the overall picture of linguistic phenomena. Examples of language studies from the network standpoint include applications in machine translation~\cite{cn1,ta1}, extractive summarization~\cite{cn1} and information retrieval. From a theoretical point of view, complex networks have been used to study patterns of language~\cite{masucci}, linguistic evolution~\cite{evolution} and to investigate the origin of fundamental properties~\cite{cancho}, such as the Zipf's law and word order. Interestingly, one of the most interesting patterns found with the study of language as a network refers to the fact that most language networks, even when modeled with different building rules, exhibit similar features, such as the so called scale-free and small-world effects~\cite{masucci}.

\Red{In the current paper, we evaluate the ability of complex networks for the Word Sense Disambiguation (WSD) task (i.e., the discrimination of which of the meanings is used in a given context for a word that has multiple meanings). Traditional approaches make use of contextual attributes of neighboring words to discriminate senses. In such methods, the context of ambiguous words is recognized with the computation of corpus-based features such as the part-of-speech of nearby words~\cite{manning}, the grammatical or syntactical relations between the word of interest and the other surrounding words~\cite{navigli}. A second approach, referred to as knowledge-based paradigm (see e.g. Ref.~\cite{lesk} and references therein), relies on the analysis of context to infer meaning with the aid of semantic resources. In this case, thesauri~\cite{roget}, ontologies~\cite{wordnet}, machine machine-readable dictionaries~\cite{mrd} and several others sources of knowledge resources~\cite{navigli,doKara} have been useful to provide semantic tips about the context of ambiguous words. In recent studies, features extracted from both corpus- and knowledge-based paradigms have been successfully integrated to yield more robust discriminative systems~\cite{doKara}. As for the automatic identification of patterns after the contextual characterization, pattern recognition techniques such as decision trees~\cite{dtrees}, Naive Bayes~\cite{manning} and neural networks~\cite{neural} have been employed.}

\Red{
While most traditional approaches disregards the connectivity patterns of words, we devised a corpus-based strategy that uses topological information extracted from networks modeling the relationship between words in the textual level~\cite{amancioEPL} and in the attribute space~\cite{thiagoHL}. Initially, the topology of ambiguous words represented as nodes in word adjacency networks was analyzed with a series of complex network metrics. Thus, each occurrence of an ambiguous word in the text is mapped into an object in the attribute space. In this space, we applied a hybrid pattern recognition which combines semantical and structural properties of ambiguous words to perform the discrimination of senses. The results suggest that word senses can be discriminated because patterns emerge not only in the adjacency networks model but also in the networks arising from the technique devised in the attribute space.} Most importantly, we show that the introduction of deterministic tourist walks over the complex network built in the attribute space improves the discrimination of senses in comparison with the same technique based on topological measurements alone.

\section{\label{sec:methodology} Methodology}

\Red{In this section, the methodology applied for discriminating word senses is discussed. The dataset is presented in Section \ref{datasetDescription} and the representation of texts as networks is explained in Section \ref{sec:rep}. Then the deterministic walk employed for characterizing word senses is discussed in Section \ref{sec:tourist}. We describe traditional classification methods in Section \ref{sec:tradClass}. Finally, the method devised in the current paper is presented in Section \ref{nbased}.}

\subsection{Dataset description} \label{datasetDescription}

\Red{
The dataset employed in the WSD task comprises $10$ words, which might assume at most four different meanings (see Table~\ref{tabsenses}). The study of the contextual topological properties of these ambiguous words was carried out on a set of $18$ books obtained from the Gutenberg\footnote{\url{gutenberg.org}} online repository. The list of books used in the current paper is presented in Table~\ref{tabbooks}.
}

\begin{table}
\centering
\Red{
    \caption{\label{tabsenses}List of 10 ambiguous words employed in the experiments and description of their respective possible senses.}
    \centering
    \begin{tabular}{|l|}
        \hline
        {\bf List of words and senses}  \\
        \hline
        bear (I) = to endure \\
        bear (II)  = mammal of the family {\it Ursidae} \\
        bear (III) = to remember (bear in mind) \\
        \hline
        jam (I) = jelly \\
        jam (II) = to compress \\
        \hline
        just (I)  = fair \\
        just (II)= only, simply \\
        just (III) = at this moment \\
        \hline
        march (I) = to trek \\
        march (II)  = third month of the calendar \\
        \hline
        rock (I)  = stone \\
        rock (II) = to sway, swing \\
        rock (III) = name (e.g. Chris Rock) \\
        \hline
        ring (I) = sound of a bell \\
        ring (II) = circle of metal \\
        \hline
        save  (I) = to rescue \\
        save  (II) = to use frugally \\
        \hline
        present (I) = now \\
        present (II)= to show \\
        \hline
        close (I) = near \\
        close (II) = shut, finish \\
        \hline
        note (I) = to take a notice \\
        note (ii) = brief informal letter \\
        \hline
    \end{tabular}
}
\end{table}

\begin{table}
    \centering
    \Red{
    \caption{\label{tabbooks}List of books (and their respective authors) employed in the experiments aiming at discriminating the meaning of ambiguous words. The year of publication is specified after the title of the book.}
    \begin{tabular}{|c|c|}
        \hline
        {\bf Title}      & {\bf Author} \\
        \hline
        Pride and Prejudice (1813) & J. Austen \\
        American Notes (1842)      & C. Dickens \\
        Coral Reefs (1842)         & C. Darwin \\
        A Tale of Two Cities (1859)& C. Dickens \\
        The Moonstone (1868)       & W. Collins \\
        Expression of Emotions (1872) & C. Darwin \\
        A Pair of Blue Eyes (1873) & T. Hardy \\
        Jude the Obscure (1895)    & T. Hardy \\
        Dracula's Guest (1897)     & B. Stoker \\
        Uncle Bernac (1897)        & A. C. Doyle \\
        The Tragedy of the Korosko (1898) & A. C. Doyle \\
        The Return of Sherlock Holmes (1903) & A. C. Doyle \\
        Tales of St. Austin's (1903) & P. G. Wodehouse \\
        The Chronicles of Clovis (1911) & H. H. Munro \\
        A Changed Man (1913) & T. Hardy \\
        Beasts and Super Beasts (1914) & H. H. Munro \\
        The Wisdom of Father Brown (1914) & G. K. Chesterton \\
        My Man Jeeves (1919) & P. G. Wodehouse \\
        \hline
    \end{tabular}
    }
\end{table}

\subsection{Representing texts as complex networks for WSD} \label{sec:rep}

Word-sense disambiguation (WSD) refers to the problem of discriminating the meaning used by a polysemous word in a particular context. The resolution of ambiguities is an essential task in the field of Natural Language Processing~\cite{manning} because other tasks (p.e., anaphora resolution, machine translation and information retrieval) depend on the accurate discrimination of senses. Traditionally, two paradigms are employed to address the problem: the deep and the shallow paradigms. While the latter employs statistical techniques to infer contextual information, the former employs knowledge bases, such as dictionaries and thesaurus. In the current paper, we have used the superficial paradigm as it usually yields the best discrimination rates~\cite{amancioEPL}.

\begin{figure}
\centering
\resizebox{0.50\textwidth}{!}
{
  \includegraphics{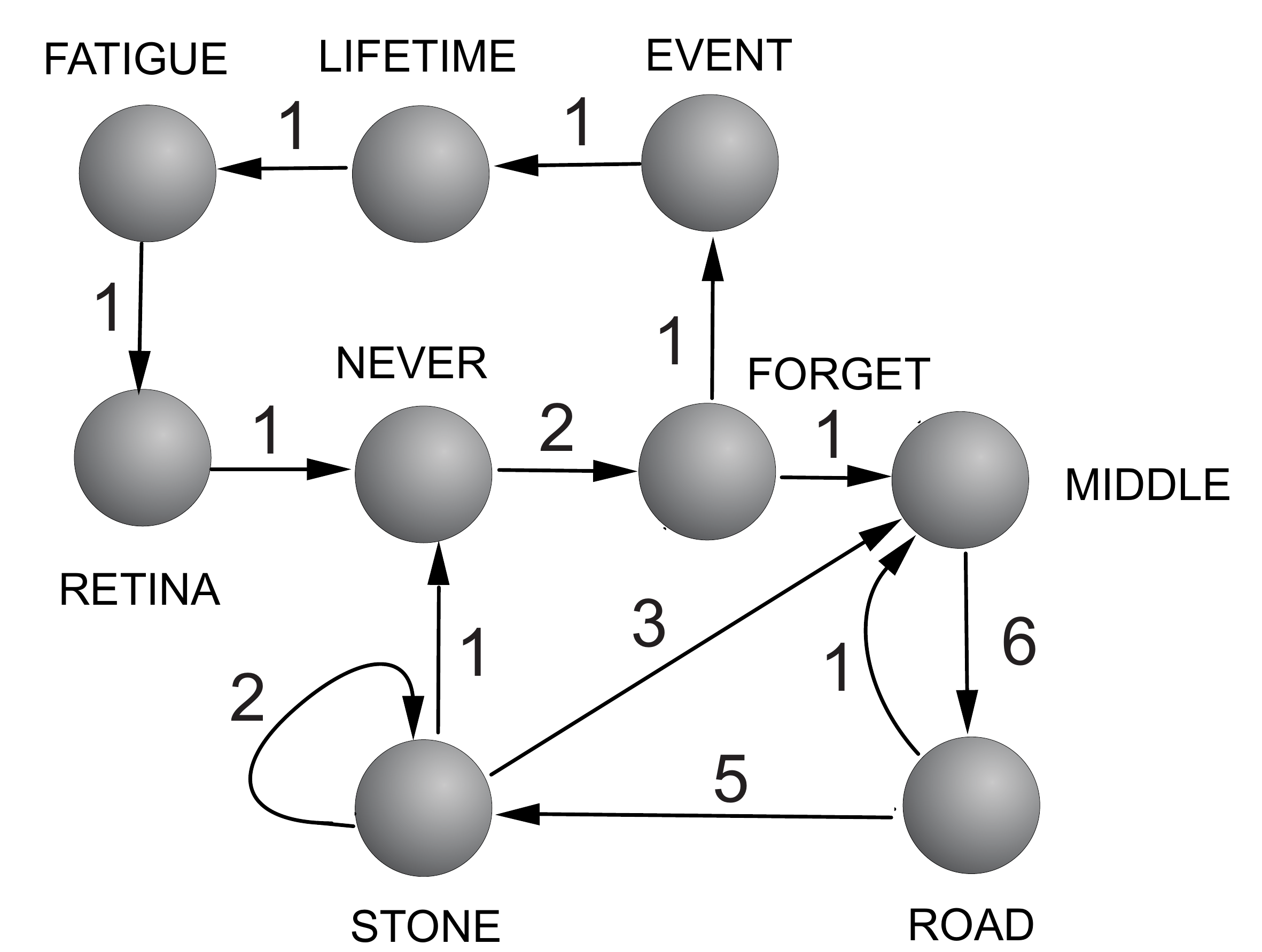}
}
%\vspace{5cm}       % Give the correct figure height in cm
\caption{\label{fig:network}\Red{Network obtained from the poem ``In the middle of the road there was a stone / there was a stone in the middle of the road / there was a stone / in the middle of the road there was a stone. / Never should I forget this event / in the lifetime of my fatigued retinas / Never should I forget that in the middle of the road / there was a stone / there was a stone in the middle of the road / in the middle of the road there was a stone.'' by Carlos Drummond de Andrade.}}
\label{img:drummond}       % Give a unique label
\end{figure}

Two superficial methods were employed in this paper. The first method, referred to as semantical approach, is based on the analysis of frequency of nearby words. More specifically, the context surrounding an ambiguous word is mathematically described by the frequency of their $\omega_n$ closest words. Particularly, we have employed $\omega_n = \{5, 20, 50\}$. \Red{In the second method, referred to as topological approach, words are represented as nodes~\cite{amancioEPL}. Edges linking nodes are created according to the order words appear in the text. As such, whenever word $i$ appears immediately before word $j$, the element $w_{ij}$ of the matrix $\mathcal{W}$ representing the network is incremented by one. Before this step, punctuation marks and stopwords (such as prepositions, articles and other high-frequency words conveying little semantic meaning) are removed. Then, the remaining words are converted to their canonical form so that words lexically distinct but semantically related are mapped into the same network node. Specifically, when modeling ambiguous words, each of its occurrence is regarded as as distinct node. An example of the pre-processing steps taken to process the poem ``In the middle of the road'' by Carlos Drummond de Andrade
(translation from Portuguese by Elizabeth Bishop) is given in Table~\ref{tab.literario}. The graphical representation of the network originated from this poem is provided in Figure~\ref{img:drummond}.
}

\begin{table*}[!ht]
\centering
\Red{
\caption{\label{tab.literario}Pre-processing steps taken to process the poem ``In the middle of the road'', by Carlos Drummond de Andrade. Initially punctuation marks and stopwords are taken away from the text. Then the remaining words are mapped into their canonical forms.}
\begin{tabular}{|l|l|}
  \hline
  \textbf{Processing Step}      & \textbf{Outcome}      \\
  \hline
  \multirow{7}{*}{\bf Original Poem}   & In the middle of the road there was a stone $/$ there was a stone \\
                        & in the middle of the road there was a stone in the middle of the  \\
                        & road there was a stone. Never should I forget this event / in the \\
                        & lifetime of my fatigued retinas / Never should I forget that in   \\
                        & the middle of the road / there was a stone / there was a stone    \\
                        & in the middle of the road / in the middle of the road there was   \\
                        & a stone. \\
  \hline
  \multirow{3}{*}{\bf Without stopwords} & middle road stone stone middle road stone middle road stone never \\
                                         & forget event lifetime fatigued retinas never forget middle road stone \\
                                         & stone middle road middle road stone \\
  \hline
  \multirow{3}{*}{\bf After Lemmatization} & middle road stone stone middle road stone middle road stone never   \\
                      & forget event lifetime fatigue retina Never forget middle road stone  \\
                      & stone middle road middle road stone                                  \\
  \hline
\end{tabular}
}
\end{table*}

The context surrounding ambiguous words is identified by the recurrence of patterns of connectivity, which are measured with the following complex networks measurements~\cite{cn1}: hierarchical degree, hierarchical clustering coefficient, average and variability of the degree of neighbors, average shortest path length and betweenness. Further details regarding the topological approach are given in the Supplementary Information\footnote{The Supplementary Information is available from \url{https://dl.dropboxusercontent.com/u/2740286/supEpjb2013.pdf}.
}

\subsection{Touristic walks} \label{sec:tourist}

A tourist walk can be conceptualized as a walker (tourist) aiming at visiting sites (data items) in a $d$-dimensional map, representing the data set. At each discrete time step, the tourist follows a simple deterministic rule: it visits the nearest site which has not been visited in the previous $\mu$ steps. In other words, the walker performs partially self-avoiding deterministic walks over the data set, where the self-avoiding factor is limited to the memory window $\mu - 1$. This quantity can be understood as a repulsive force emanating from the sites in this memory window, which prevents the walker from visiting them in this interval (refractory time). Therefore, it is prohibited that a trajectory to intersect itself inside this memory window. In spite of being a simple rule, it has been shown that this movement dynamic possesses complex behavior when $\mu > 1$ \cite{Lima2001}.

\begin{figure}
\centering
\resizebox{0.4\textwidth}{!}
{
  \includegraphics{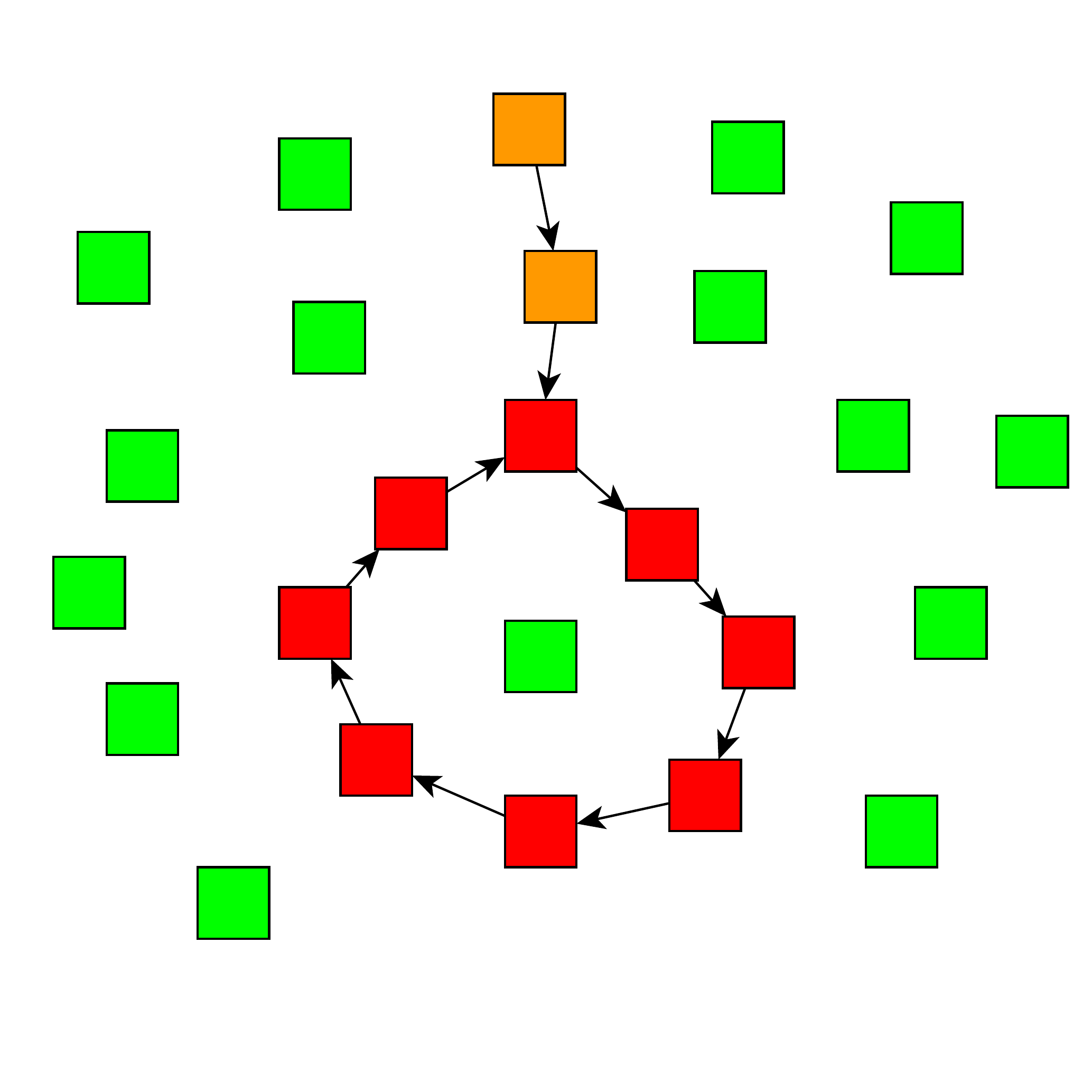}
}
%\vspace{5cm}       % Give the correct figure height in cm
\caption{Illustration of a tourist walk with $\mu = 1$. The green dots represent unvisited sites. The orange and red sites indicate visited sites, where the first color displays the transient part of the tourist walk and the second color, the cycle of the walk.}
\label{img:tourist-walk-schematic}       % Give a unique label
\end{figure}

Each tourist walk can be decomposed in two terms: (i) the initial \emph{transient part} of length $t$ and (ii) a \emph{cycle} (attractor) with period $c$. Figure \ref{img:tourist-walk-schematic} shows an illustration of a tourist walk with $\mu = 1$, i.e., the walker always goes to the nearest neighbor. In this case, one can see that the transient length is $t=2$ and the cycle length $c=8$.

\subsection{Traditional classification} \label{sec:tradClass}

\Red{
The traditional bag-of-words~\cite{manning} technique employed for the WSD task is exemplified in this section with three low-level algorithms, namely Naive Bayes, kNN and C4.5. For all the three pattern recognition techniques, each content word surrounding the polysemous word is taken as a single attribute~\cite{manning}.
}

\subsubsection{Naive Bayes}

\Red{
 The Naive Bayes technique is based on the following rule:
\begin{description}
  \item[{\bf Bayesian optimal decision rule}]: choose sense $s'$ if the condition
  \begin{equation}
  P(s'|c) > P(s_k|c)
  \end{equation}
  holds for each $s_k \neq s'$, where $P(s_k|c)$ is the probability of the sense $s_k$ to occur in the context $c$.
\end{description}
}

\begin{figure}
\centering
\resizebox{0.4\textwidth}{!}
{
  \includegraphics{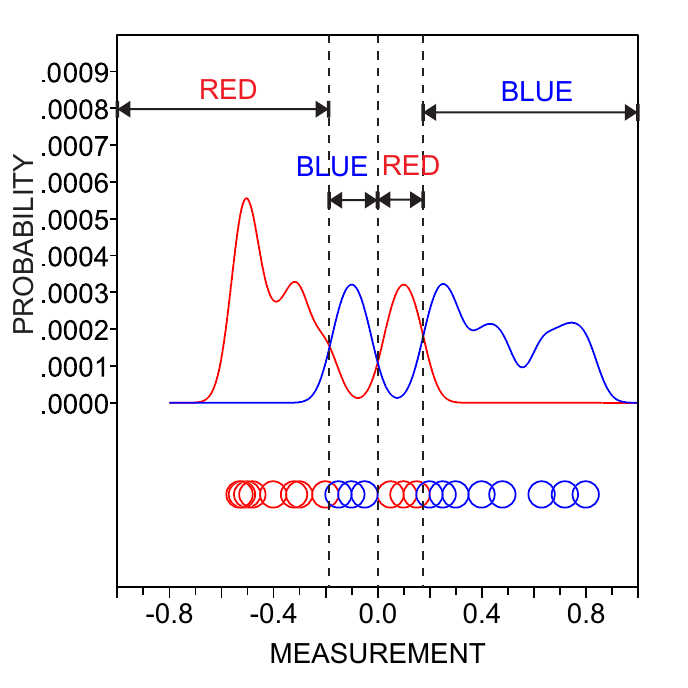}
}
%\vspace{5cm}
% Give the correct figure height in cm
\caption{\Red{Example of decision based on the Naive Bayes algorithm. Any word satisfying the condition $P(v_j|\textrm{``red''}) > P(v_j|\textrm{``blue''})$ will be classified as $s'=$``red''. Analogously, $P(v_j|\textrm{``blue''}) > P(v_j|\textrm{``red''})$ determines the boundaries of the ``blue'' sense decision region.}
}
\label{bayes_1}       % Give a unique label
\end{figure}

\Red{
Usually, the values of $P(s_k|c)$ are unknown. Nevertheless, they can be recovered straightforwardly from the Bayes's theorem:}
\begin{equation}
    P(s_k|c) = \frac{ P(c|s_k) }{P(c)} P(s_k).
\end{equation}
\Red{
$P(s_k)$ is referred to as {\it a priori} probability of $s_k$, i.e., the probability of $s_k$ occur in any context.
The evidence concerning the context is given by the term  $P(c|s_k)$. Because we are tackling an instantiation of a discrimination task, $P(c)$ can be disregard for it is remains constant for each sense $s_k$. As such, the most likely sense $s'$ is given by:
\begin{eqnarray}
s' &=& \arg\max_{s_k} P(s_k|c)      \nonumber \\
   &=& \arg\max_{s_k} \frac{ P(c|s_k) }{P(c)} P(s_k) \nonumber \\
   &=& \arg\max_{s_k} P(c|s_k) P(s_k) \nonumber \\
   &=& \arg\max_{s_k} [\log P(c|s_k) + \log P(s_k)].
\end{eqnarray}
%
%Here the logaritm was taken in order to simplify next computations.
%
Upon employing the hypothesis of attribute independence and considering that a neighbor word $v_j$ represents the context, i.e. $c = \{v_j | v_j \in c \}$, consider the following assumption:
\begin{description}
  \item[{\bf Naive Bayes hypothesis}]:
  \begin{equation}
  P(c|s_k) = P( \{v_j | v_j \in c \} | s_k \} = \prod_{v_j \in c} P(v_j|s_k).
  \end{equation}
\end{description}
Note that if the Naive Bayes hypothesis is assumed to be true all the organization of the contextual words is disregarded. In addition, due to the independence assumption, the presence of an specific word in the context does not affect the probability of occurrence of other words. Finally, the most likely sense of the polysemous word is computed as:
\begin{equation} \label{dec.bayesiana}
    s' = \arg\max_{s_k} [\log P(s_k) + \sum_{v_j \in c} \log P(v_j|s_k)].
\end{equation}
}
\Red{
In order to illustrate the process of deciding the correct sense of an ambiguous word with the Naive Bayes technique, consider Figure \ref{bayes_1}. The position of each circle represents the proximity of a given word $v_j$ to one of the two senses: $s=$``red'' or $s=$``blue''. Since both senses occur with the same frequency the effective term accounting for the decision is $\log P(v_j|s_k)$ in equation \ref{dec.bayesiana}. In the current paper, the likelihood $P(v_j|s_k)$ is estimated with the Parzen windowing strategy~\cite{duda}. The decision boundaries are then established according to equation \ref{dec.bayesiana}. For example, any new instance lying in the interval between -0.19 and 0.00 will be classified as $s'=$``blue'' because this sense is the most likely sense in that region. In the same way, the interval lying between 0.00 and 0.18 is governed by the ``red'' sense because $P(v_j|\textrm{``red''}) > P(v_j|\textrm{``blue''})$ in this particular interval.
}

\subsubsection{$k$-nearest neighbors (kNN)}
\begin{figure}
\centering
\resizebox{0.4\textwidth}{!}
{
  \includegraphics{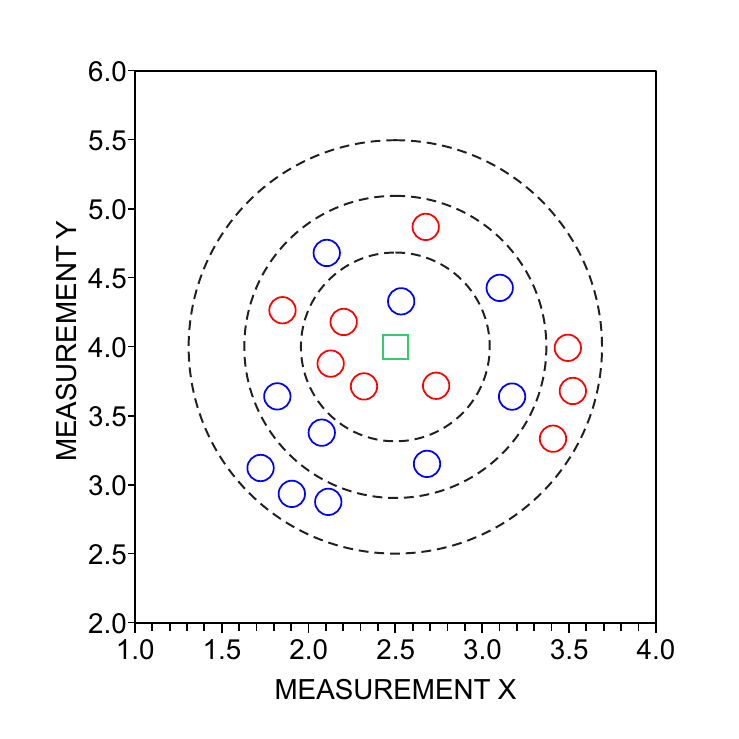}
}
%\vspace{5cm}
% Give the correct figure height in cm
\caption{\Red{Example of classification with the $k$-nearest neighbor technique. Note that the decision relies on a voting process performed over the $k$ of nearest neighbors of the ambiguous word, which is represent by the green square. In the experiments we employed $k = 1$ along with the Euclidean distance.}}
\label{knn}       % Give a unique label
\end{figure}

\Red{
In this technique, each sense $s$ of a given ambiguous word is attributed according to a voting process performed over the set of the $k$ nearest neighbors (computed according to a similarity measure) comprising the instances labeled previously. If most of the $k$ nearest neighbors of the ambiguous word were labeled with sense $s'$,  then sense $s'$ is attributed to the word. To illustrate how the algorithm performs the classification, consider Figure \ref{knn}. Suppose that we aim at classifying the sense of an polysemous   word illustrated by the green square in Figure \ref{knn}. If the decision is taken with $k=5$ (see the innermost dotted circle), the correct sense would be $s'=$``red'' because four out of the five nearest neighbors belong to the ``red'' class.  When the decision is performed with $k=13$ (see the central dotted circle) the majority class is the blue one. As a result, $s'=$``blue'' is the sense associated. In a similar fashion, when $k=19$ (see the outermost dotted circle) the sense associated to the word remains $s'=$``blue''.
}

\subsubsection{Decision trees}

\begin{figure}
\centering
\resizebox{0.4\textwidth}{!}
{
  \includegraphics{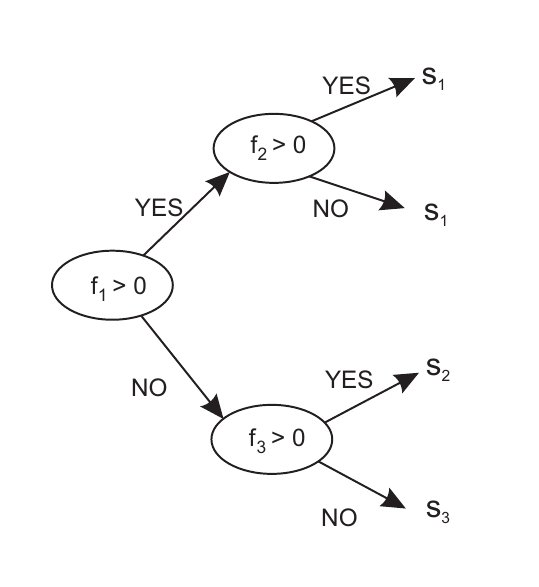}
}
%\vspace{5cm}
% Give the correct figure height in cm
\caption{\Red{Example of decision tree employed to decide the correct sense of an ambiguous word, described by the following attributes: $f_1 = -0.23$, $f_2 = +0.29$ and $f_3 = +0.38$.
The first test leads to the edge labeled as ``NO'' (because $f_1 > 0$) and the second test leads to the edge ``YES'' (because $f_3 > 0$). Therefore,
the sense $s_3$ is associated to the ambiguous word. }}
\label{fig:arvore}       % Give a unique label
\end{figure}

\Red{The decision tree algorithm employed in the current paper is the so called C4.5~\cite{quinlan}. The tree used for the decision comprises internal and leaf nodes, where the former represent attributes and the latter represent classes (in our case, word senses).
Edges represent conditional clauses, such as ``IF (attribute $\geq$ threshold)'' or ``IF (attribute $\leq$ threshold)''. To classify an unknown instance, one walks on the tree following the rules applied to the attribute related to the current internal node until one reaches a leaf node. At this point, the class assigned is the class stored in the leaf node. To construct a decision tree, the algorithm C4.5 chooses an attribute $f_i$ and a threshold $L$ that best separates the dataset~\cite{quinlan}. In this case, the criterion used to distinguish items is the information gain~\cite{duda}, which represents the difference in entropy $\mathcal{H}$ before and after using the attribute to separate the data. More specifically, the information gain related to the attribute $f_i$ in the training set~\cite{thiagoHL} $\mathcal{S}_{\textrm{tr}}$ is given by:
\begin{equation}
    \Omega(\mathcal{S}_{\textrm{tr}},f_i) = \mathcal{H}(\mathcal{S}_{\textrm{tr}}) -  \mathcal{H}(\mathcal{\mathcal{S}_{\textrm{tr}}}|f_i).
\end{equation}
Thus, the attribute with largest information gain receives the highest priority to perform the discrimination of senses. The process is reiterated recursively for the two children nodes  of the current node. When separation is complete, i.e. all instances resulting from the separation belongs to a unique class, the process stops and a leaf node is created to store the corresponding class. The classification process is exemplified  in Figure \ref{fig:arvore}.} Further details regarding this method is provided in Ref.~\cite{quinlan}.

\subsection{Network-based classification} \label{nbased}

In a typical supervised data classification task, a data set $X$ is decomposed into two disjoint sets: the training set $X_{\mathrm{training}}$, where the labels $y_{i} \in \mathcal{L} = \{L_1,\ldots,L_n\}$ are known, and the test set $X_{\mathrm{test}}$, in which the labels are unknown. Each instance of $X_{\mathrm{training}}$ or $X_{\mathrm{test}}$ is a $d$-dimensional vector-based data item \cite{thiagoSSL,thiagoUP}. Usually, the generalization power of the classifier is quantified by measuring the accuracy rate achieved when it is applied to the test set. The training set is used in the \emph{training phase} to induce the hypothesis of the classifier, while the test set is utilized in the \emph{classification phase} to predict new unseen instances. In the following, these two phases are discussed.

In the training phase, the data in the training set are mapped into a graph ${G}$ using a network formation technique $g: X_{training} \mapsto G = \langle {V}, {E} \rangle$, where ${V} = \{1,\ldots,V\}$ is the set of vertices and ${E}$ is the set of edges. Each vertex in ${V}$ represents a training instance in ${X}_{{training}}$. As it will be described later, the pattern formation of the classes will be extracted by using the complex topological features of this networked representation.

The edges in ${E}$ are created using a combination of the $\epsilon_r$ and the $k$-nearest neighbors ($\kappa$NN) graph formation techniques. In the original versions, the $\epsilon_r$ technique creates a link between two vertices if they are within a distance $\epsilon$, while the $\kappa$NN sets up a link between vertices $i$ and $j$ if $i$ is one of the $k$ nearest neighbors of $j$ or vice versa. Both approaches have their limitations when sparsity or density is a concern. For sparse regions, the $\kappa$NN forces a vertex to connect to its $k$ nearest vertices, even if they are far apart. In this scenario, one can say that the neighborhood of this vertex would contain dissimilar points. Equivalently, improper $\epsilon$ values could result in disconnected components, sub-graphs, or isolated singleton vertices.

\begin{figure}
\centering
\resizebox{0.5\textwidth}{!}
{
  \includegraphics{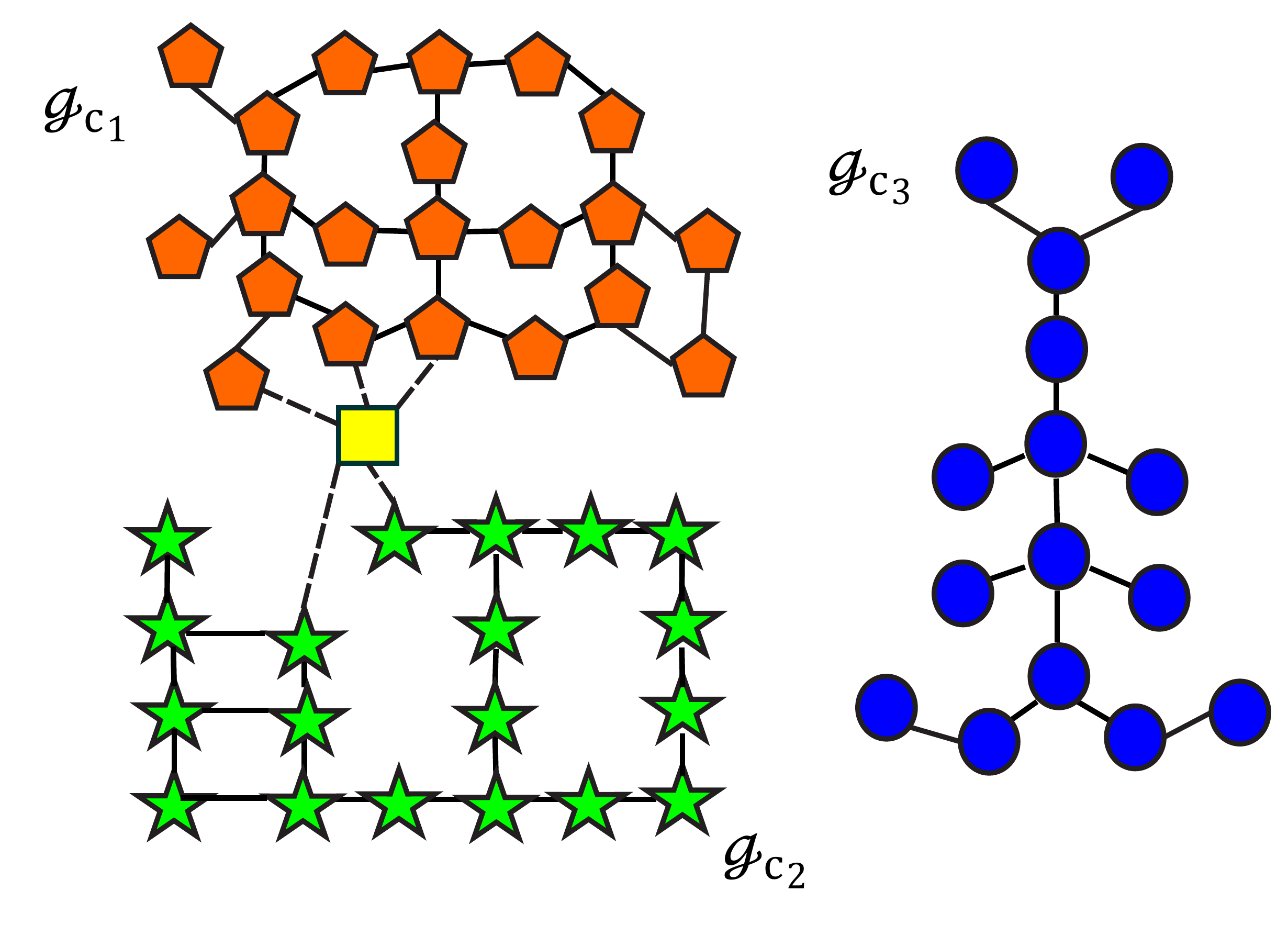}
}
%\vspace{5cm}
% Give the correct figure height in cm
\caption{Schematic of the constructed network. At the end of the training phase, each class is represented by a graph component. In the classification phase, a new test instance (yellow or ``square-shaped'' data) is temporarily inserted into the graph.}
\label{schematic-test}       % Give a unique label
\end{figure}

The network is constructed using these two traditional graph formation techniques in a combined form. The neighborhood of a vertex $x_{i}$ is given by $N(x_i) = \epsilon_r(x_i, y_{x_i})$, if $|\epsilon_r$($x_{i}$, $y_{x_{i}}| > k$. Otherwise, $N(x_i) = \kappa(x_i, y_{x_i})$, where $y_{x_{i}}$ denotes the class label of the training instance $x_{i}$, \mbox{$\epsilon_r(x_{i}, y_{x_{i}})$} returns the set $\{x_{j}, j \in \mathcal{V} : d(x_{i},x_{j}) < \epsilon \wedge y_{x_{i}} = y_{x_{j}}\}$, and $\kappa(x_{i}, y_{x_{i}})$ returns the set containing the $k$ nearest vertices of the same class as $x_{i}$. Note that the $\epsilon_r$ technique is used for dense regions ($|\epsilon_r(x_{i})| > k$), while the $\kappa$NN is employed for sparse regions. With this mechanism, it is expected that each class will have a unique and single graph component. For example, Fig. \ref{schematic-test} displays a schematic of how the network looks like for a three-class problem when the training phase has been completed. One can see that each class holds a representative component.

In the classification phase, the unlabeled data items in the $\mathbf{X}_{\mathrm{test}}$ are presented to the classifier one by one. In contrast to the training phase, the class labels of the test instances are unknown. This test instance is inserted also using a mixture of the $\epsilon$-radius and $\kappa$NN. In this case, the labels are not taken into consideration. Once the data item is inserted, each class analyzes, in isolation, its impact on the respective class component using the complex topological features of it. In the high level model, each class retains an isolated graph component. Each of these components calculate the changes that occur in its pattern formation with the insertion of this test instance. If slight or no changes occur, then it is said that the test instance is in compliance with that class pattern. As a result, the high level classifier yields a great membership value for that test instance on that class. Conversely, if these changes dramatically modify the class pattern, then the high level classifier produces a small membership value on that class. These changes are quantified via network measures, each of which numerically translating the organization of the component from a local to global fashion. For this end, the touristic walks are employed in a novel manner. Again, in Fig. \ref{schematic-test}, a test instance denoted by the yellow or ``square-shaped'' data item is temporarily inserted into the network (dashed lines). Due to its insertion, the class components become altered: $\mathcal{G}_{C_{1}}^{'}$, $\mathcal{G}_{C_{2}}^{'}$, and $\mathcal{G}_{C_{3}}^{'}$. It may occur that some class components do not share any links with this test instance. In the figure, this happens with $\mathcal{G}_{C_{3}}^{'}$. In this case, we say that test instance do not comply to the pattern formation of the class component. For the components that share at least a link ($\mathcal{G}_{C_{1}}^{'}$ and $\mathcal{G}_{C_{2}}^{'}$), each of it calculates considers only the edges of the test instance that are linked to it in order to measure the compliance. For instance, when we check the compliance of the test instance with the component $\mathcal{G}_{C_{1}}^{'}$, the connections from the test instance to the component $\mathcal{G}_{C_{2}}^{'}$ are ignored, and vice versa.

Simultaneously to the prediction yielded by the high level classifier, a low level classifier also predicts the membership of the same test instance. The way it predicts depends on the choice of the low level classifier. In the end, the predictions produced by both classifiers are combined via a linear combination to derive the prediction of the high level framework (meta-learning). Once the test instance gets classified, it is either discarded or incorporated to the training set with the corresponding predicted label. In the second case, the classifier must be retrained. Note that, in any of the two situations, each class is still represented by a single graph component.

\subsubsection{Hybrid high level classification}
\label{Classification-Technique}

The hybrid high level classifier $M$ consists of a convex combination of two terms: (i) a low level classifier, for instance, a decision tree, SVM, or a $k$NN classifier; and (ii) a high level classifier, which is responsible for classifying a test instance according to its pattern formation with the data. Mathematically, the membership of the test instance $x_{i} \in \mathbf{X}_{\mathrm{test}}$ with respect to the class $j \in \mathcal{L}$, here written as $M^{(j)}_{i}$, is given by:

\begin{equation}
    M^{(j)}_{i} = (1 - \lambda)L^{(j)}_{i} + \lambda H^{(j)}_{i}
    \label{eq:def-classification}
\end{equation}

\noindent where $L^{(j)}_{i} \in [0,1]$ denotes the membership of the test instance $x_{i}$ on class $j$ produced by an arbitrary traditional (low level) classifier; $H^{(j)}_{i} \in [0,1]$ indicates the same membership information yielded by a high level classifier; $\lambda \in [0,1]$ is the \emph{compliance term}, which plays the role of counterbalancing the classification decision supplied by both low and high level classifiers. Whenever $L^{(j)}_{i} = 1$ and $H^{(j)}_{i} = 1$, we may deduce that the $i$th data item
carries all the characteristics of class $j$. On the other hand, whenever $L^{(j)}_{i} = 0$ and $H^{(j)}_{i} = 0$, we may infer that the $i$th data item does not present any similarities nor complies to the pattern formation of class $j$. Values in-between these two extremes
lead to natural uncertainness in the classification process and are found in the majority of times during a classification
task. Also, observe that, when $\lambda = 0$, (\ref{eq:def-classification}) reduces to a common low level classifier.

A test instance receives the label from the class $j$ that maximizes (\ref{eq:def-classification}). Mathematically, the estimated label of the test instance $x_{i}$, $\hat{y}_{x_{i}}$, is given by:

\begin{equation}
    \hat{y}_{x_{i}} = \mathrm{arg}_{j \in {L}}  {\mathrm{max}\mbox{ }} M^{(j)}_{i}. %\operatorname{}
    \label{eq:classification-equation}
\end{equation}

Motivated by the intrinsic ability to describe topological structures among the data items, the high level term of the hybrid classifier makes use of networks to infer decisions. The only restriction imposed on the nature of the network is that each class must be an isolated subgraph (component). In order to extract the complex properties of the network, two network measures derived from the tourist walks dynamics with different memory lengths $\mu$ will be utilized, namely the transition length $T(\mu)$ and the cycle length $C(\mu)$ of a walk. By varying the memory length of a walk, one is able to capture from local to global complex features of the class component. Mathematically, the membership of the test instance $x_{i} \in \mathbf{X}_{\mathrm{test}}$ with respect to the class $j \in \mathcal{L}$ yielded by the high level classifier, here written as $H^{(j)}_{i}$, is given by:

\begin{equation}
    H^{(j)}_{i} = \frac{\sum_{\mu=0}^{\mu_{c}}{\left[\alpha_{t}(1 - T_{i}^{(j)}(\mu)) + \alpha_{c}(1 - C_{i}^{(j)}(\mu)) \right]} }{\sum_{g\in \mathcal{L}}{\sum_{\mu=0}^{\mu_{c}}{\left[\alpha_{t}(1 - T_{i}^{(g)}(\mu)) + \alpha_{c}(1 - C_{i}^{(g)}(\mu)) \right]}}}
    \label{eq:def-C-term}
\end{equation}

\noindent where $\mu_{c}$ is a critical value that indicates the maximum memory length of the tourist walks,
$\alpha_{t},\alpha_{c} \in [0,1], \alpha_{t} + \alpha_{c} = 1$, are user-controllable coefficients that indicate the influence of each network measure in the
process of classification, $T_{i}^{(j)}(\mu)$ and $C_{i}^{(j)}(\mu)$ are functions that depend on the transient and cycle
lengths, respectively, of the tourist walk applied to the $i$th data item with regard to the class $j$. These functions are
responsible for providing an estimative whether the data item $i$ under analysis possesses the same patterns of the component
$j$ or not. The denominator in (\ref{eq:def-C-term}) has been introduced solely for normalization matters.

Regarding $T_{i}^{(j)}(\mu)$ and $C_{i}^{(j)}(\mu)$, they are given by the following expressions:

\begin{equation}
    %\begin{split}
    T_{i}^{(j)}(\mu) = \Delta t^{(j)}_{i}(\mu) p^{(j)} \ \ \ \ \ \\
    C_{i}^{(j)}(\mu) = \Delta c^{(j)}_{i}(\mu) p^{(j)}
    %\end{split}
    \label{eq:f-function-def}
\end{equation}

\noindent where $\Delta t^{(j)}_{i}(u), \Delta c^{(j)}_{i}(u) \in [0,1]$ are the variations of the transient and cycle lengths that occur on the component representing class $j$ if $i$ joins it and $p^{(j)} \in [0,1]$ is the proportion of data items pertaining to class $j$.

Remembering that each class has a component representing itself, the strategy to check the pattern compliance of a test instance is to examine whether its insertion causes a great variation of the network measures representing the class component. In other words, if there is a small change in the network measures, the test instance is in compliance with all the other data items that comprise that class component, i.e., it follows the same pattern as the original members of that class. On the other hand, if its insertion is responsible for a significant variation of the component's network measures, then probably the test instance may not belong to that class. This is exactly the behavior that (\ref{eq:def-C-term}) together with (\ref{eq:f-function-def}) propose, since a small variation of $T(\mu)$ or $C(\mu)$ causes a large membership value $H$, and vice versa.

Next, we explain how to compute $\Delta t^{(j)}_{i}(\mu)$ and $\Delta c^{(j)}_{i}(\mu)$ that appear in (\ref{eq:f-function-def}). Firstly, we need to numerically quantify the transient and cycle lengths of a component. Since the tourist walks are strongly dependent on the starting vertices, for a fixed $\mu$, we perform tourist walks initiating from each one of the vertices that are members of a class component. The transient and cycle lengths of the $j$th component, $\langle t^{(j)} \rangle$ and $\langle c^{(j)} \rangle$, are simply given by the average transient and cycle lengths of all its vertices, respectively. In order to estimate the variation of the component network measures, consider that $x_{i} \in \mathbf{X}_{\mathrm{test}}$ is a test instance. In relation to an arbitrary class $j$, we virtually insert $x_{i}$ into component $j$ using the  plain $\epsilon$-radius technique, and recalculate the new average transient and cycle lengths of this component. We denote these new values as $\langle {t'}^{(j)} \rangle$ and $\langle {c'}^{(j)} \rangle$, respectively. This procedure is performed for all classes $j \in \mathcal{L}$. It may occur that some classes $u \in \mathcal{L}$ will not share any connections with the test instance $x_{i}$. Using this approach, $\langle t^{(k)} \rangle = \langle {t'}^{(k)} \rangle$ and $\langle c^{(k)} \rangle = \langle {c'}^{(k)} \rangle$, which is undesirable, since this configuration would state that $x_{i}$ complies perfectly with class $u$. In order to overcome this problem, a simple post-processing is necessary: For all components $u \in \mathcal{L}$ that do not share at least $1$ link with $x_{i}$, we deliberately set $\langle {t'}^{(j)} \rangle$ and $\langle {c'}^{(j)} \rangle$ to a high value. This high value must be greater than the largest variation that occurs in a component which shares a link with the data item under analysis. One may interpret this post-processing as a way to state that $x_{i}$ do not share any pattern formation with class $u$, since it is not even connected to it.

With all this information at hand, we are able to calculate $\Delta t^{(j)}_{i}(\mu)$ and $\Delta c^{(j)}_{i}(\mu),\forall j \in \mathcal{L}$, as follows:

\begin{eqnarray}
    %\begin{split}
    \Delta t^{(j)}_{i}(\mu) &=& \frac{|\langle {t'}^{(j)} \rangle - \langle {t}^{(j)} \rangle|}{\sum_{u \in \mathcal{L}}{|\langle {t'}^{(u)} \rangle - \langle {t}^{(u)} \rangle|}}\\
    \Delta c^{(j)}_{i}(\mu) &=& \frac{|\langle {c'}^{(j)} \rangle - \langle {c}^{(j)} \rangle|}{\sum_{u \in \mathcal{L}}{|\langle {c'}^{(u)} \rangle - \langle {c}^{(u)} \rangle|}}
    %\end{split}
    \label{eq:deltaG-def}
\end{eqnarray}

The network-based high level classifier quantifies the variations of the transient and cycle lengths of tourist walks with limited memory $\mu$ that occur in the class components when a test instance artificially joins each of them in isolation. According to (\ref{eq:def-C-term}), this procedure is performed for several values of the memory length $\mu$, ranging from $0$ (memoryless) to a critical value $\mu_{c}$. This is done in order to capture complex patterns of each of the representative class components in a local to global fashion. When $\mu$ is small, the walks tend to possess a small transient and cycle parts, so that the walker does not wander far away from the starting vertex. In this way, the walking mechanism is responsible for capturing the local structures of the class component. On the other hand, when $\mu$ increases, the walker is compelled to venture deep into the component, possibly very far away from its starting vertex. In this case, the walking process is responsible for capturing the global features of the component. In summary, the fundamental idea of the high level classifier is to make use of a mixture of local and global features of the class components by means of a combination of tourist walks with different values of $\mu$.

\section{Results and discussion}

First, the methodology is applied to an artificial database in order to better understand its functionality. Afterwards, the WSD problem is analyzed. The discussion of the observed results is given below.

\subsection{High Level Applied to a Toy Database}

\begin{figure*}
\centering
\resizebox{1\textwidth}{!}
{
  \includegraphics{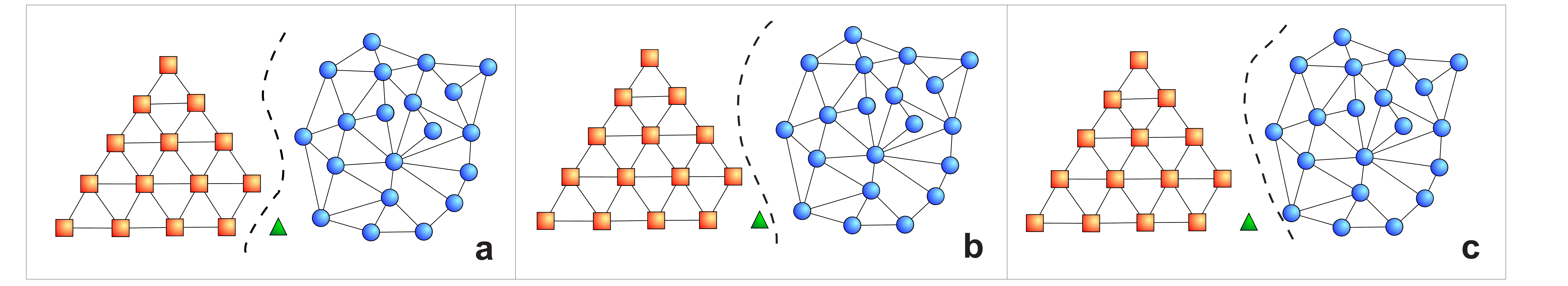}
}
%\vspace{5cm}       % Give the correct figure height in cm
\caption{An artificial data set containing a structured and a non-structured class. The behavior of the decision boundaries varies as $\lambda$ increases. Decision boundaries correspond to (a) $\lambda = 0$; (b) $\lambda = 0.5$; and (c) $\lambda = 0.8$.}
\label{fig:Regular-Lattice-lambda0}       % Give a unique label
\end{figure*}

In order to better understand the dynamics of the model, consider the toy data set depicted in Fig. \ref{fig:Regular-Lattice-lambda0}(a),
where there are two classes: the orange or ``square'' ($14$ vertices) and the blue or ``circle" ($20$ vertices) classes. This example serves as a gist of how the hybrid classifier draws its decisions. In the training and classification phases, we employ $\kappa = 3$ and $\epsilon = 0.02$ for the network construction. The fuzzy SVM~\cite{svmref} with RBF kernel ($C=40$ and $\gamma=2^{2}$) is adopted for the low level classifier.  By inspection of the figure, the orange or ``square" class displays a well-defined pattern: a pyramidal structure, whereas the blue or ``circle" class does not indicate any well-established patterns. Here, the goal is to classify the green or ``triangle'' data item using only the information of the training set. Figures \ref{fig:Regular-Lattice-lambda0}(a), \ref{fig:Regular-Lattice-lambda0}(b), and \ref{fig:Regular-Lattice-lambda0}(c) exhibit the decision boundaries of the two classes when $\lambda = 0$, $\lambda = 0.5$, and $\lambda = 0.8$, respectively. When $\lambda = 0$, only the SVM prediction is used by the hybrid technique. In this case, one can see that the triangle-shaped data item is not correctly classified. Notice that the decision boundaries are pushed near the blue or ``circle'' class by virtue of the large amount of blue or ``square'' items in the vicinity. Now, when $\lambda = 0.5$, the SVM and the high level classifier predictions are utilized in the same intensity. In this situation, the decision boundaries are dragged toward the orange or ``square'' class, because of the strong pattern that it exhibits. We can think this phenomenon as being a clash between the two decision boundaries: as $\lambda$ increases, the more structured class tends to possess more decision power, and, consequently, is able to reduce the effective area of the competing class. For example, when $\lambda = 0.8$, the organizational features of the orange or ``square'' class are so salient that its effective area invades the high density region of the blue or ``circle'' class. In the two former cases, the hybrid high level technique can successfully classify the triangle-shaped data item.

\subsection{High-Level applied to word sense disambiguation}

The high-level technique was applied to the problem of discriminating senses of $10$ polysemous words. \Red{We used the database described in Section~\ref{datasetDescription}}, which is available online\footnote{\Red{The database of features extracted from the set of ambiguous words is available from \url{dl.dropbox.com/u/2740286/database_words.zip}}}. For the characterization of senses, two paradigms were employed: the traditional based on the recurrence of neighbors (see Table \ref{tab.1}) and the one based on the structure of the word adjacency network (see Table \ref{tab.2}). In both cases, we can see that the best results occur when $\lambda > 0$. This means that structural patterns in the attribute space emerge both on the semantical and topological characterization. In particular, the strongest dependence (i.e., the lowest $p$-values) between word senses and structural organization in the word adjacency networks was observed for the words {\it rock} and {\it close}. For the semantic approach, the strongest dependence was found for the words {\it ring} and {\it rock}.

\begin{table*}
    \centering
    \caption{\label{tab.1} Semantic approach for discriminating senses of ambiguous words. Senses were characterized according to frequency of the $n=5$ neighbors of the ambiguous word~\cite{amancioEPL} and the discrimination of senses was performed with low (kNN, C4.5 and Bayes) and high level classifiers. Acc. Rate represents the accuracy rate obtained with an evaluation based on the 10-fold cross-validation technique~\cite{validation}. The p-value refers to the likelihood of obtaining the same accuracy rate with an random classifier (see Ref.~\cite{amancioEPL} for details). Note that the high level technique always outperforms the traditional low level classification.}
    \begin{tabular}{|c|c|c|c|c|c|c|}
        \hline
        Approach      & \multicolumn{3}{|c|}{Low Level Classification} & \multicolumn{3}{|c|}{High Level Classification}               \\
        \hline
        \textbf{Word} & \textbf{Algorithm} &\textbf{Acc. Rate} & \textbf{p-value} & \textbf{Acc. Rate} & \textbf{p-value} & \textbf{Best $\lambda$}  \\
        \hline
                       & kNN   & 79.5 \%  & $4.6 \times 10^{-1}$  & 83.3 \% & $1.8 \times 10^{-1}$ & 0.20 \\
        save           & C4.5  & 78.9 \%  & $5.6 \times 10^{-1}$  & 85.8 \% & $7.3 \times 10^{-2}$ & 0.30 \\
                       & Bayes & 76.6 \%  & $7.5 \times 10^{-1}$  & 83.3 \% & $1.8 \times 10^{-1}$ & 0.35 \\
        \hline
                       & kNN   & 82.6 \%  & $1.3 \times 10^{-1}$ & 87.9 \% & $7.6 \times 10^{-3}$  & 0.25\\
        note           & C4.5  & 79.5 \%  & $3.7 \times 10^{-1}$ & 85.0 \% & $5.3 \times 10^{-2}$  & 0.25\\
                       & Bayes & 78.3 \%  & $4.6 \times 10^{-1}$ & 85.0 \% & $5.3 \times 10^{-2}$  & 0.30\\
        \hline
                       & kNN   & 82.8 \%  & $8.1 \times 10^{-3}$  & 90.5 \% & $3.3 \times 10^{-4}$ & 0.35\\
        march          & C4.5  & 82.8 \%  & $8.1 \times 10^{-3}$  & 90.5 \% & $3.3 \times 10^{-4}$ & 0.30\\
                       & Bayes & 62.5 \%  & $4.2 \times 10^{-1}$  & 71.4 \% & $1.5 \times 10^{-1}$ & 0.30\\
        \hline
                       & kNN   & 62.9 \%  & $7.0 \times 10^{-1}$ & 68.8 \% & $1.0 \times 10^{-2}$  & 0.20\\
        present        & C4.5  & 57.0 \%  & $9.8 \times 10^{-1}$ & 60.1 \% & $8.9 \times 10^{-1}$  & 0.10\\
                       & Bayes & 60.2 \%  & $8.9 \times 10^{-1}$ & 64.4 \% & $5.3 \times 10^{-1}$  & 0.15\\
        \hline
                       & kNN   & 76.5 \%  & $1.6 \times 10^{-2}$ & 90.0 \% & $4.6 \times 10^{-2}$ & 0.40\\
        jam            & C4.5  & 76.5 \%  & $4.6 \times 10^{-2}$ & 90.0 \% & $4.6 \times 10^{-2}$ & 0.45\\
                       & Bayes & 90.0 \%  & $4.6 \times 10^{-2}$ & 90.0 \% & $4.6 \times 10^{-2}$ & 0.35\\
        \hline
                       & kNN   & 80.7 \%  & $7.2 \times 10^{-3}$ & 86.4 \% & $3.4 \times 10^{-5}$ & 0.25\\
        ring           & C4.5  & 84.4 \%  & $2.8 \times 10^{-4}$ & 89.8 \% & $7.0 \times 10^{-7}$ & 0.20\\
                       & Bayes & 83.5 \%  & $7.0 \times 10^{-4}$ & 87.0 \% & $3.4 \times 10^{-5}$ & 0.15\\
        \hline
                       & kNN   & 47.4 \%  & $1.1 \times 10^{-1}$ & 59.1 \% & $6.8 \times 10^{-5}$ & 0.50 \\
        just           & C4.5  & 38.0 \%  & $8.0 \times 10^{-1}$ & 48.5 \% & $5.5 \times 10^{-2}$ & 0.40 \\
                       & Bayes & 52.6 \%  & $5.9 \times 10^{-3}$ & 63.3 \% & $1.0 \times 10^{-6}$ & 0.45 \\
        \hline
                       & kNN   & 48.0 \%  & $6.2 \times 10^{-1}$ & 53.7 \% & $2.3 \times 10^{-1}$ & 0.20 \\
        bear           & C4.5  & 62.0 \%  & $8.0 \times 10^{-3}$ & 65.4 \% & $1.1 \times 10^{-3}$ & 0.15 \\
                       & Bayes & 48.0 \%  & $6.2 \times 10^{-1}$ & 54.9 \% & $1.3 \times 10^{-1}$ & 0.30 \\
        \hline
                       & kNN   & 69.7 \%  & $3.6 \times 10^{-4}$ & 69.7 \%  & $3.6 \times 10^{-4}$ & 0.00 \\
        rock           & C4.5  & 68.5 \%  & $7.4 \times 10^{-4}$ & 70.0 \%  & $1.7 \times 10^{-4}$ & 0.10 \\
                       & Bayes & 77.5 \%  & $1.0 \times 10^{-7}$ & 77.5 \%  & $1.0 \times 10^{-7}$ & 0.00 \\
        \hline
                       & kNN   & 64.8 \%  & $9.1 \times 10^{-1}$  & 70.5 \% & $2.1 \times 10^{-1}$ & 0.25 \\
        close          & C4.5  & 71.9 \%  & $1.0 \times 10^{-2}$  & 76.1 \% & $1.7 \times 10^{-3}$ & 0.15 \\
                       & Bayes & 75.3 \%  & $3.9 \times 10^{-3}$  & 80.0 \% & $3.7 \times 10^{-6}$ & 0.15 \\
        \hline
        \end{tabular}
    \end{table*}

\begin{table*}
\centering
\caption{\label{tab.2} Structural approach for discriminating senses of ambiguous words. Senses were characterized according to topological CN measurements~\cite{amancioEPL} and the discrimination of senses was performed with low (kNN, C4.5 and Bayes) and high level classifiers.  Note that the high level technique always outperforms the traditional low level classification.}
    \begin{tabular}{|c|c|c|c|c|c|c|}
    \hline
    Approach      & Low Level  & \multicolumn{2}{|c|}{Low Level Classification} & \multicolumn{3}{|c|}{High Level Classification}                \\
    \hline
    \textbf{Word} & \textbf{Algorithm} & \textbf{Acc. Rate} & \textbf{p-value} & \textbf{Acc. Rate} & \textbf{p-value} & \textbf{Best $\lambda$}  \\
    \hline
                    & kNN  & 87.6 \%  &  $2.1 \times 10^{-2}$  & 90.9 \% &  $1.6 \times 10^{-3}$ & 0.15 \\
    save            & C4.5 & 79.8 \%  &  $4.5 \times 10^{-1}$  & 86.0 \% &  $4.1 \times 10^{-2}$ & 0.25 \\
                    & Bayes& 83.1 \%  &  $1.8 \times 10^{-1}$  & 86.0 \% &  $4.1 \times 10^{-2}$ & 0.10 \\
    \hline
                    & kNN  & 84.5 \%  &  $2.6 \times 10^{-1}$ & 86.1 \% & $1.1 \times 10^{-1}$ & 0.10 \\
    note            & C4.5 & 78.4 \%  &  $8.2 \times 10^{-1}$ & 81.4 \% & $5.6 \times 10^{-1}$ & 0.10 \\
                    & Bayes& 78.4 \%  &  $8.2 \times 10^{-1}$ & 81.4 \% & $5.6 \times 10^{-1}$ & 0.10 \\
    \hline
                    & kNN  & 87.0 \%  &  $1.9 \times 10^{-3}$ & 89.3 \% & $3.3 \times 10^{-4}$ & 0.15 \\
    march           & C4.5 & 60.9 \%  &  $4.2 \times 10^{-1}$ & 73.4 \% & $6.8 \times 10^{-2}$ & 0.40 \\
                    & Bayes& 73.9 \%  &  $6.8 \times 10^{-2}$ & 77.9 \% & $2.6 \times 10^{-2}$ & 0.25 \\
    \hline
                    & kNN  & 71.1 \%  &  $2.2 \times 10^{-3}$ & 73.6 \% & $2.7 \times 10^{-3}$ & 0.30  \\
    present         & C4.5 & 64.7 \%  &  $4.7 \times 10^{-1}$ & 67.8 \% & $1.7 \times 10^{-1}$ & 0.25  \\
                    & Bayes& 73.9 \%  &  $1.7 \times 10^{-3}$ & 77.8 \% & $2.8 \times 10^{-5}$ & 0.40  \\
    \hline
                    & kNN & 100.0 \% &  $6.0 \times 10^{-3}$ & 100.0 \% &  $6.0 \times 10^{-3}$ & 0.00  \\
    jam             & C4.5 & 80.0 \% &  $1.7 \times 10^{-1}$ & 85.6 \%  &  $4.6 \times 10^{-2}$  & 0.20  \\
                    & Bayes& 90.0 \% &  $4.6 \times 10^{-2}$ & 92.8 \%  &  $4.6 \times 10^{-2}$ & 0.10  \\
    \hline
                    & kNN   & 84.6 \%  &  $2.8 \times 10^{-4}$ & 92.4 \% & $3.3 \times 10^{-8}$  & 0.35 \\
    ring            & C4.5  & 71.4 \%  &  $2.8 \times 10^{-1}$ & 78.3 \% & $2.5 \times 10^{-1}$  & 0.30 \\
                    & Bayes & 69.2 \%  &  $4.6 \times 10^{-1}$ & 75.7 \% & $6.9 \times 10^{-2}$  & 0.25 \\
    \hline
                    & kNN   & 51.3 \%  &  $1.6 \times 10^{-2}$ & 60.0 \% & $3.1 \times 10^{-5}$ & 0.45 \\
    just            & C4.5  & 47.0 \%  &  $1.1 \times 10^{-1}$ & 55.8 \% & $1.0 \times 10^{-3}$ & 0.35 \\
                    & Bayes & 46.2 \%  &  $1.5 \times 10^{-1}$ & 53.3 \% & $5.9 \times 10^{-3}$ & 0.30 \\
    \hline
                    & kNN  & 62.0 \%  &  $8.0 \times 10^{-3}$ & 67.2 \% & $2.6 \times 10^{-4}$ & 0.25 \\
    bear            & C4.5 & 62.0 \%  &  $8.0 \times 10^{-3}$ & 68.9 \% & $1.1 \times 10^{-4}$ & 0.30 \\
                    & Bayes& 60.9 \%  &  $1.4 \times 10^{-2}$ & 67.0 \% & $2.6 \times 10^{-4}$ & 0.25 \\
    \hline
                    & kNN  & 78.4 \%  &  $3.2 \times 10^{-8}$ & 82.1 \% & $1.9 \times 10^{-10}$ & 0.15 \\
    rock            & C4.5 & 79.3 \%  &  $9.7 \times 10^{-9}$ & 84.5 \% & $2.3 \times 10^{-12}$ & 0.15 \\
                    & Bayes& 70.3 \%  &  $1.8 \times 10^{-4}$ & 73.3 \% & $1.5 \times 10^{-5}$ & 0.10 \\
    \hline
                    & kNN   & 72.2 \%  &  $8.0 \times 10^{-2}$ & 81.8 \% & $6.0 \times 10^{-8}$ & 0.35 \\
    close           & C4.5  & 68.7 \%  &  $4.7 \times 10^{-1}$ & 80.1 \% & $1.9 \times 10^{-6}$ & 0.40 \\
                    & Bayes & 68.7 \%  &  $4.7 \times 10^{-1}$ & 79.4 \% & $6.8 \times 10^{-6}$ & 0.35 \\
    \hline
    \end{tabular}
\end{table*}

\begin{figure*}
\centering
\resizebox{0.8\textwidth}{!}
{
  \includegraphics{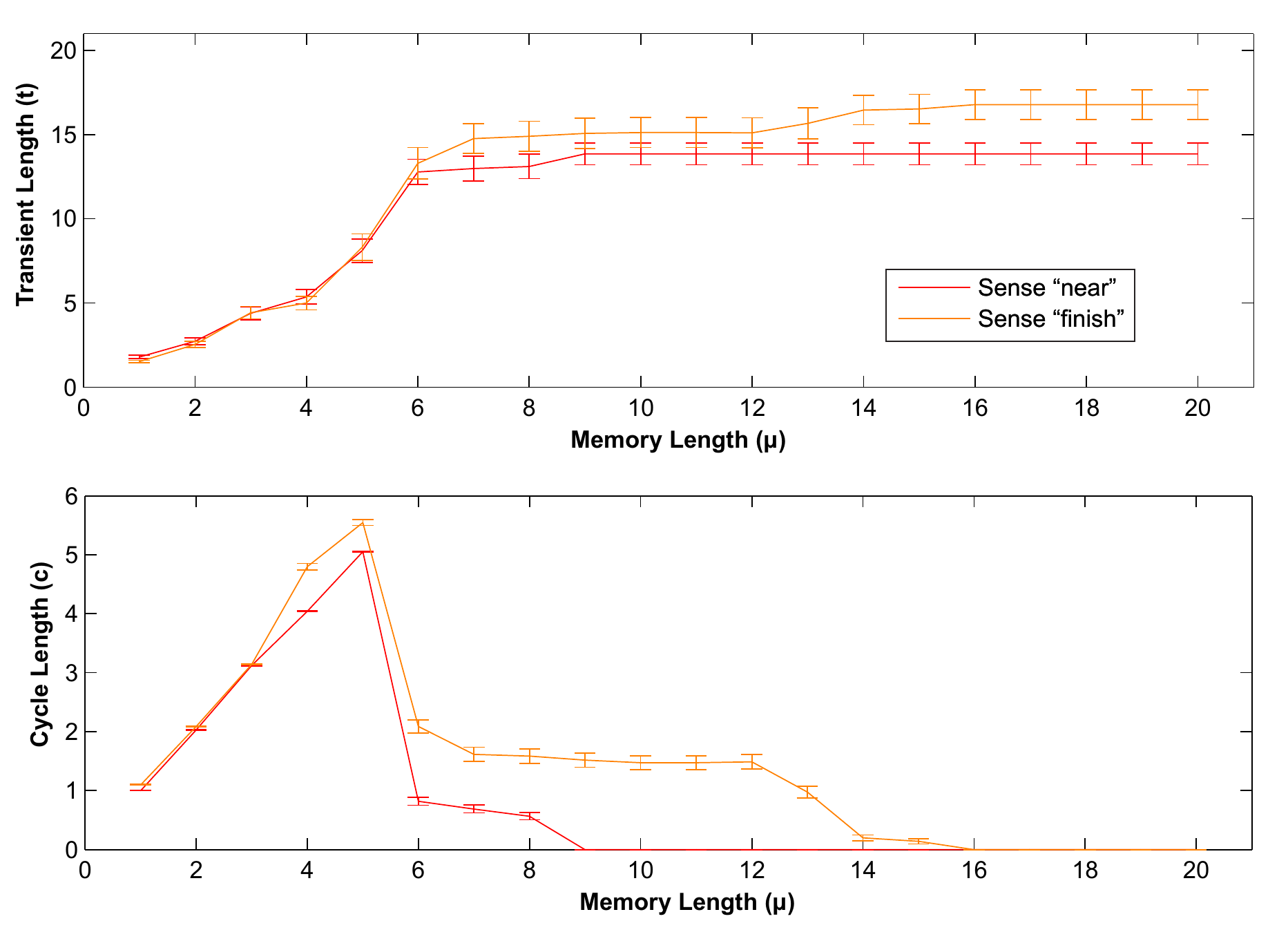}
}
%\vspace{5cm}       % Give the correct figure height in cm
\caption{\Red{Analysis of variability of $t$ and  $c$ with $\mu$ for two senses of the word ``close''. $t$ and $c$ are useful to discriminate senses for $\mu \geq 6$. The time to reach the steady state (bottom panel) suggests that distinct senses are characterized by distinct patterns of complexity.}}
\label{fig:memoryclose}       % Give a unique label
\end{figure*}

\begin{figure*}
\centering
\resizebox{0.8\textwidth}{!}
{
  \includegraphics{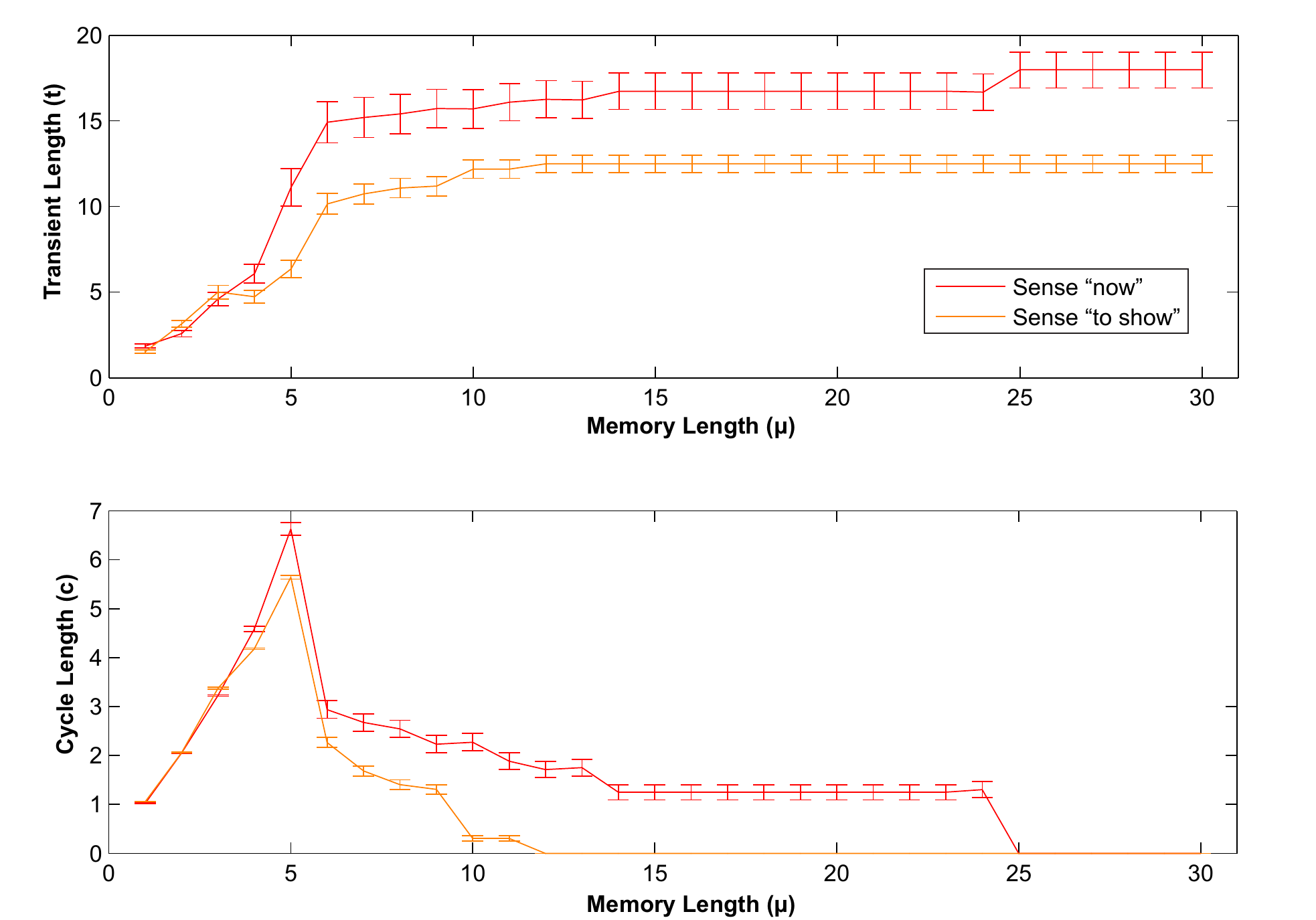}
}
%\vspace{5cm}       % Give the correct figure height in cm
\caption{\Red{Analysis of variability of $t$ and $c$ with $\mu$ for two senses of the word ``present''. $t$ and $c$ are useful to discriminate senses for $\mu \geq 4$ and $\mu \geq 6$, respectively. The time to reach the steady state (bottom panel) suggests that distinct senses are characterized by distinct patterns of complexity.}}
\label{fig:memorypresent}       % Give a unique label
\end{figure*}

One can wonder the reason behind retrieving a walk with
such complex behavior to solve this kind of problem in detriment
to simply using well-known complex network measurements,
such as assortativity, clustering coefficient, or average degree. This was devised on account of the way that the tourist walks artlessly supply structural information of the network in a local to global fashion by simple ranging the size of the memory of the tourist. As we increment the memory of the tourist, it gradually captures more global information of the network, since the cycle and transient lengths tend to vary more in relation to small values of $\mu$. In contrast to that, each complex network measurement captures, in a static manner, a specific network's characteristic. For instance, the degree strictly captures local organization of each vertex; the clustering coefficient captures quasi-local relations, since it investigates the neighborhood's structure, in search of triangles; and the assortativity aims at exclusively unveiling global information of each network component. Therefore, the idea of gradually capturing the network characteristics of the network from a local to global fashion ceases to exist when one utilizes pre-made network measurements. As a result, this makes room for an evident gap that links the measurements that analyze completely local and global peculiarities of the network. One could argue that other new measurements could be plugged into the high order of learning term of the classifier. This would solve the problem at the cost of creating unnecessary complexity to the model, as well as spending a great amount of time to devise these specific network measurements. As we can see, when we utilize the tourist walks with varying memory lengths, this gap is filled in a concise and clear manner by just incrementing $\mu$. \Red{This effect is illustrated in Figures \ref{fig:memoryclose} and \ref{fig:memorypresent} for the words ``close'' and ``present'', respectively\footnote{\Red{The relationship between $t$ and $c$ with $\mu$ for the other words is presented in the SI.}}. As one can observe, in the WSD task a distinct behavior arises both in the evolution of the $t$ and $c$ for specific word senses. In the case of the word ``close'', the transient lengths discriminate the two possible senses for a memory length $\mu \geq 6$. With regard to the word ``present'', a discrimination of senses occurs for $\mu \geq 5$. The dependence of the cycle length $c$ with $\mu$ also seems to be sense-specific. Remarkably, distinct senses are characterized by distinct times for $c$ to reach the steady state, i.e., the state where there is no variation in response to the increase of $\mu$. These results suggests that the complexity of each component representing a word sense can be grasped with the introduction of further memory lengths and then employed to improve the discrimination of polysemous words.
}

\section{Conclusion}

Patterns of structural organization are often used in the process of human learning. Nevertheless, automatic pattern recognition techniques oftentimes disregard these patterns. For this reason, in the current study, we have applied a recognition technique based on organizational patterns specifically to the problem of word senses recognition in a given context. Through the semantic and topological characterization of the words, we have noticed that the discrimination is improved with the high-level technique. These results suggest that there are organizational patterns in the resulting space, regardless of the paradigm employed to characterize the word senses.  Most importantly, we have found that the characterization via touristic walks yielded better discrimination rates when compared to the characterization based only in complex networks measurements, such as clustering coefficient and betweenness~\cite{amancioEPL}. A reason for this important phenomenon was discussed. This finding also suggests that this kind of characterization may be useful in other related applications. At last, we have found that both syntactic and semantic factors play a key role in identifying the meaning of the words, for the best approach (semantical or topological) depended from word to word.

Future works could study how one can automatically choose the best value for the compliance term. We also intend to apply the same methodology devised here with others classifying schemes~\cite{alneu,ivan} and in other classification tasks~\cite{rezende}, such as the problems of disambiguating names in citation networks~\cite{newman,amancionovo,citations}.

\section*{Acknowledgments}
TCS (2009/12329-1) and DRA (2010/00927-9) acknowledge the financial support from FAPESP.


\end{document}